\documentclass{article}

\pdfoutput=1

\usepackage{arxiv}

\usepackage[utf8]{inputenc} 
\usepackage[T1]{fontenc}    
\usepackage{url}            
\usepackage{booktabs}       
\usepackage{amsfonts}       
\usepackage{nicefrac}       
\usepackage{microtype}      
\usepackage{lipsum}		
\usepackage{graphicx}
\usepackage{doi}
\usepackage{multirow}
\usepackage{tabularx}
\usepackage{amsmath}
\usepackage{gensymb}

\title{Estimating Depth of Monocular Panoramic Image with Teacher-Student Model Fusing Equirectangular and Spherical Representations}

\author{Jingguo Liu$^{1}$, Yijun Xu$^{1}$, Shigang Li$^{2}$, Jianfeng Li$^{1}$\thanks{Corresponding author}
\\
$^{1}$Southwest University, Chongqing, China\\
$^{2}$Hiroshima City University, Hiroshima, Japan\\
{\tt\small popqlee@swu.edu.cn}\\
}

\renewcommand{\shorttitle}{\textit{arXiv} Template}

\begin{document}
\maketitle

\begin{abstract}
Disconnectivity and distortion are the two problems which must be coped with when processing 360 degrees equirectangular images. In this paper, we propose a method of estimating the depth of monocular panoramic image with a teacher-student model fusing equirectangular and spherical representations. In contrast with the existing methods fusing an equirectangular representation with a cube map representation or tangent representation, a spherical representation is a better choice because a sampling on a sphere is more uniform and can also cope with distortion more effectively. In this processing, a novel spherical convolution kernel computing with sampling points on a sphere is developed to extract features from the spherical representation, and then, a Segmentation Feature Fusion(SFF) methodology is utilized to combine the features with ones extracted from the equirectangular representation.
In contrast with the existing methods using a teacher-student model to obtain a lighter model of depth estimation, we use a teacher-student model to learn the latent features of depth images. This results in a trained model which estimates the depth map of an equirectangular image using not only the feature maps extracted from an input equirectangular image but also the distilled knowledge learnt from the ground truth of depth map of a training set. In experiments, the proposed method is tested on several well-known 360 monocular depth estimation benchmark datasets, and outperforms the existing methods for the most evaluation indexes.
\end{abstract}


\section{Introduction}
\label{sec:intro}

Wider field of view means richer visual information.
Estimating the depth from a single 360\degree panoramic image is an interesting topic, and until now a lot of researches have reported on it \cite{22,23,acdnet,slice,17,20,21}.
Since a 360\degree panoramic image is usually represented as an Equi-Rectangular Projection(ERP)\cite{11,12}, this problem is formulated as the estimation of depth from a single ERP image concretely.

However, when a 360\degree panoramic image is represented as an ERP image, the problems of disconnectivity and distortion arise.
While the disconnectivity can be solved easily by padding the left side using the right side image, how to coping with the distortion is tricky.
In the existing methods, combining a cubemap representation \cite{10} with an ERP image is used cope with this problems \cite{13,14}.
In comparison with the distortion increasing greatly as approaching to poles of an ERP image, a cubemap representation is made up of six square perspective images.

Although a cubemap representation of a 360\degree panoramic image can improve the distortion of an ERP image effectively, it has its own limitations. 
First, since a cubemap representation is made up of six square perspective images, padding operations is necessary when carrying out convolution on the boundaries of each perspective image.
Next, theoretically, a cubemap representation is not a ideal one for a 360\degree panoramic image to cope with image distortion because a perspective image has its own distortion.

Similarly, tangent representation\cite{9} is proposed to use to cope with the distortion. Tangent representation represents the panoramic image with any number of perspective images. However, due to the large number of views, there is a significant amount of redundancy in many regions, and the fusion processing of these repetitive regions will introduce new issues\cite{16}.

It is known that an ideal representation for a 360\degree panoramic image is a spherical image because the distortion of a scene object does not change with its position on a sphere.
This isotropic property of a spherical image makes it superior to other representations.
Additionally, on a sphere the problem of disconnectivity is eliminated completely.
In this paper, we estimate the depth of a monocular panoramic image by fusing a spherical representation with an ERP image.
A spherical convolution method is also developed, which enables a spherical convolution is carried out on sphere directly.
Moreover, the feature maps extracted by the spherical convolution is fused with those extracted from the ERP image to achieve better performance than the existing methods. 

Additionally, the existing methods of estimating depth of monocular panoramic image use the known ground truth of depth map in loss function to update the parameters of neural network during the back-propagation process.
On the other hand, three dimensional structure of environments has its own inherent characteristics, especially for indoor environment having ceilings, floors and walls.

Based on this idea, we design a teacher-student model to learn the inherent cues of depth images of training set.

In this paper, we train an encoder-decoder structure with depth image input and depth image output to extract the inherent characteristics of panoramic depth images first, and then using this pretrained model as the teacher model to supervise the student network learning. The experimental results show that the accuracy of depth estimation is improved.

To evaluate the proposed approach, we conducted experiments on the 3D60\cite{3d60}, Matterport3D\cite{m3d}, and Stanford2D3D\cite{s2d3d} datasets. The results demonstrate that our method surpasses existing approaches on the Matterport3D\cite{m3d} and Stanford2D3D\cite{s2d3d} datasets and achieves competitive performance on the 3D60\cite{3d60} dataset. In summary, the contributions of this paper are as follows:
\begin{itemize}
    \item In contrast with the existing methods fusing an ERP representation with a cubemap representation or a tangent representation, a Segmentation Feature Fusion(SFF) methodology is designed to combine spherical representation with the equirectangular representation to improve the performance of depth estimation. 
    \item To realize the spherical representation, we design a new spherical kernel to carry out spherical convolution on a sphere, which solves the problems of disconnectivity and distortion of an ERP image effectively. 
    \item We propose an encoder-decoder network to exploit the inherent cues of depth images of training set and supervise the backbone network learning in a distilled knowledge way. Our proposed teacher-student model is different from the existing methods which only use depth map as ground truth in the loss function of the network output at training phase.
\end{itemize}
\section{Related Work}
\label{sec:related}
\subsection{Monocular 360 depth estimation}
Monocular 360 depth estimation is an extension of monocular depth estimation that focuses on predicting depth information in a 360-degree panoramic view by utilizing a single image as input. For example, \cite{20} explored the spherical view synthesis to learn monocular 360-degree depth via a self-supervised method. \cite{21} builds a two-stage pipeline for omnidirectional monocular depth estimation.
\cite{17} predicts the depth directly on the spherical mesh without projection preprocessing and achieved a good results.
To address the spherical distortion in ERP images, \cite{22} employed deformable convolution to adapt the sampling grids in response to geometric distortions within panoramic images. Moreover, \cite{acdnet} adaptively combines convolution kernels with varying dilations to expand the receptive field.

\cite{23} devised a distortion-aware deformable convolution filter for testing purposes, a filter that can be trained using conventional perspective images. Differently, \cite{slice} represents the scene as compact vertical slices of a sphere and predict depth with convolution layers.
These methods have demonstrated the feasibility of applying convolution directly on ERP images to eliminate distortions.

\begin{figure*}[t]
  \centering
  \includegraphics[width=0.92\linewidth]{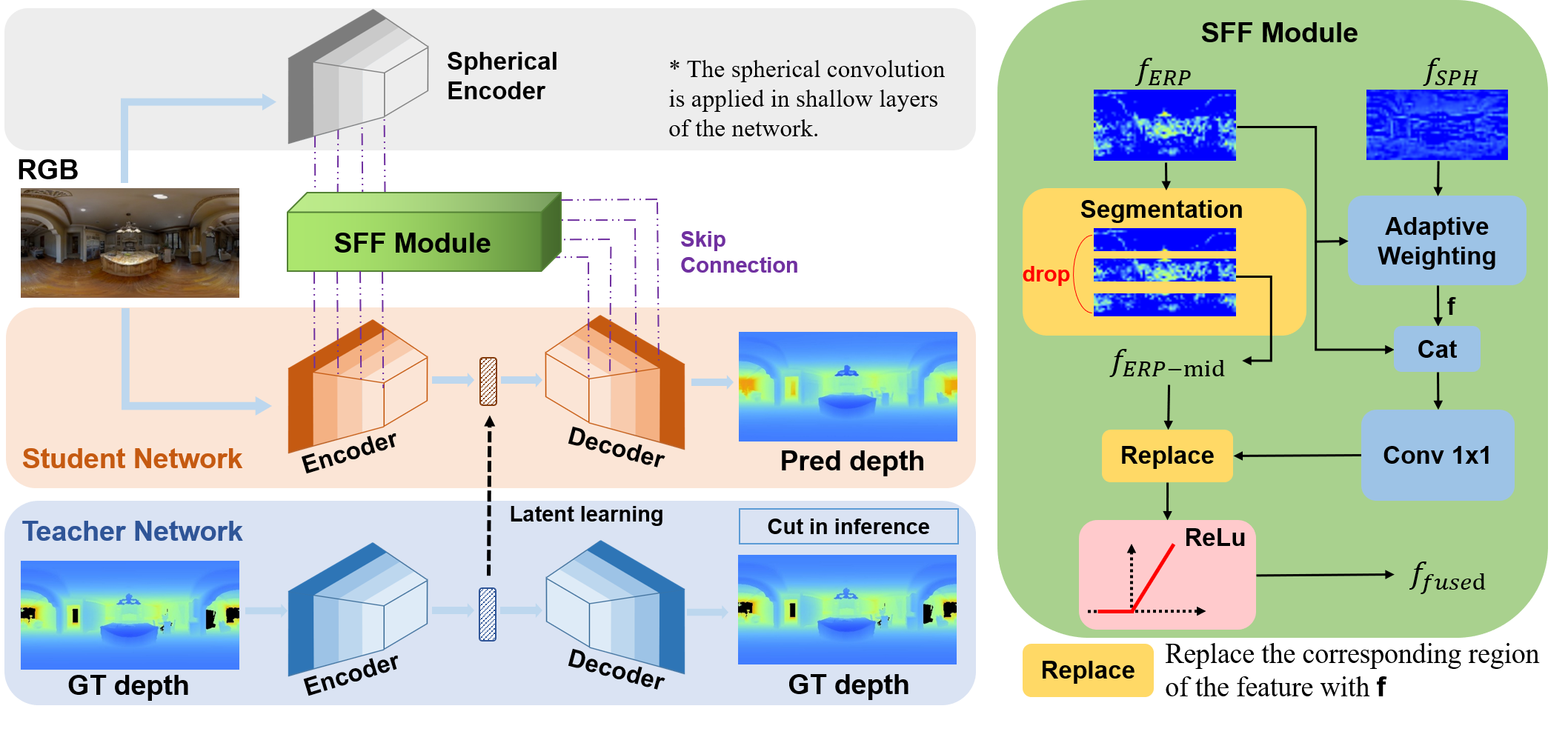}
  \caption{Overview of our network}
  \label{Fig.2}
\end{figure*}
Recently, there has been a growing interest in utilizing fusion-based approaches to cope with the distortion. \cite{13} proposed to effectively combines the cubemap and ERP features from both the encoder and decoder stages. Furthermore, \cite{14} proposed a new framework for fusing features from different projections: ERP and Cubemap
and demonstrate that the ERP features are more important for final ERP format depth prediction tasks.
Differently, \cite{24} designed a Cubemap Vision Transformers to extract distortion-free global features from the panorama and fuse them at multiple scales. For tangent patches based fusion methods, \cite{15} proposed to estimate the depth from tangent patches and fuse the tangent patches to an ERP image. \cite{PanelNet} introduce Local2Global Transformer, which aggregates local information within a panel and panel-wise global context. overhead. \cite{panoformer} introduced a panoramic transformer designed to exploit tangent patches within the spherical domain. 
\cite{16} combined CNN and transformer to learn the holistic contextual information from the ERP and tangent patches and adopts a classification model for depth value prediction.
\cite{EGformer} propose an equirectangular geometry-biased transformer.

In contrast with the closest researches\cite{13,14,16}, we propose a novel approach that fuses ERP and spherical representations. This integration can mitigate the defectives caused by ERP representation most effectively.

\subsection{Spherical convolution}
Spherical convolution is characterized by capturing and preserving the spatial information from panoramic images. Recently, \cite{26} designed a Kernel Transformer to transfer the convolution kernels from perspective images to ERP images. \cite{28} proposed to use spherical convolution to deal with the problem of weight sharing failure caused by video projection distortion. \cite{29} employed spherical convolution to distill spatial-temporal 360 information. cite{30} presented a spherical CNN that constructed by representing the sphere as a graph, and utilized the graph-based representation to define the standard CNN operations. These methods have provided evidence for the effectiveness of spherical convolutions in processing information from panoramic image. \cite{OSRT} design a distortion-aware Transformer to modulate ERP distortions continuously and self-adaptively. \cite{Spherephd} proposed to utilizes a spherical polyhedron to represent omni-directional views to minimizes the variance of the spatial resolving power on the sphere surface.

\subsection{Knowledge distillation}
Knowledge distillation aims to enable the student model to mimic the behavior and performance of the teacher model. Knowledge distillation is first proposed by \cite{32}.

It is worth noting that \cite{33} proposed that semantically similar inputs tend to elicit similar activation patterns in a trained network. Moreover, \cite{34} demonstrated that knowledge distillation can be a powerful tool for reducing the size of large models without compromising their performance. These methods provide ample evidence of the effectiveness of knowledge distillation, in the field of deep learning. Moreover, some methods\cite{Focal,Wavelet,Variational} have proved that the teacher-student model learning at the latent feature level is a feasible and effective approach.

In this paper we propose a network to exploit the inherent cues of depth images of training set and supervise the backbone network learning in a distilled knowledge way.
Our proposed teacher-student model is different from the existing methods\cite{111,222,31} which only use depth map as ground truth in the loss function of the network output at training phase.
\section{Proposed Methods}
\subsection{Overview}
The proposed framework introduces a novel approach for monocular panoramic depth estimation. \ref{Fig.2} shows the framework, which incorporates an ERP-based teacher-student model and employs spherical convolution for distortion elimination.

In our network, an ERP image serves as the input, and the predicted depth is output. The encoder utilizes a ConvNeXt-base pretrained model\cite{convnet} to extract features from the input with channel numbers of [128, 256, 512, 1024]. Similarly, the spherical convolution encoder applies the proposed spherical convolution method to extract distortion-free high-dimensional features of corresponding sizes and channels in shallow networks. Besides, a skip connection structure (similar to \cite{14}) 

is applied to enhance the interaction between the encoder and decoder and enrich the high-dimensional information of the image. Following an encoder-decoder architecture, the teacher network takes the ground truth depth image as input. In contrast to conventional depth estimation methods, our framework harnesses the benefits of knowledge distillation networks by employing a teacher network trained with ground truth to extract the inherent characteristics of the depth image. 
In decoder stage, the interpolation-based upsampling method is used to upsample the obtained features. Notably, We utilize a sub-pixel convolution\cite{sup} for final upsampling layers, which can minimize the impact of excessive manual factors on the results and enhance the spatial details.

\subsection{Spherical convolution}
\subsubsection{Spherical kernel}
One crucial aspect of performing convolution on a sphere is setting up the appropriate convolution kernel. Different with planar convolutions, convolution on a spherical surface possesses a distinctive characteristic: the kernels, whether rectangular or Gaussian, do not undergo translation but instead rotation on the sphere. Therefore, the problem of defective rotational invariance of convolution kernels on the sphere cannot be ignored.
\begin{figure}[t]
  \centering
  \includegraphics[width=0.95\linewidth]{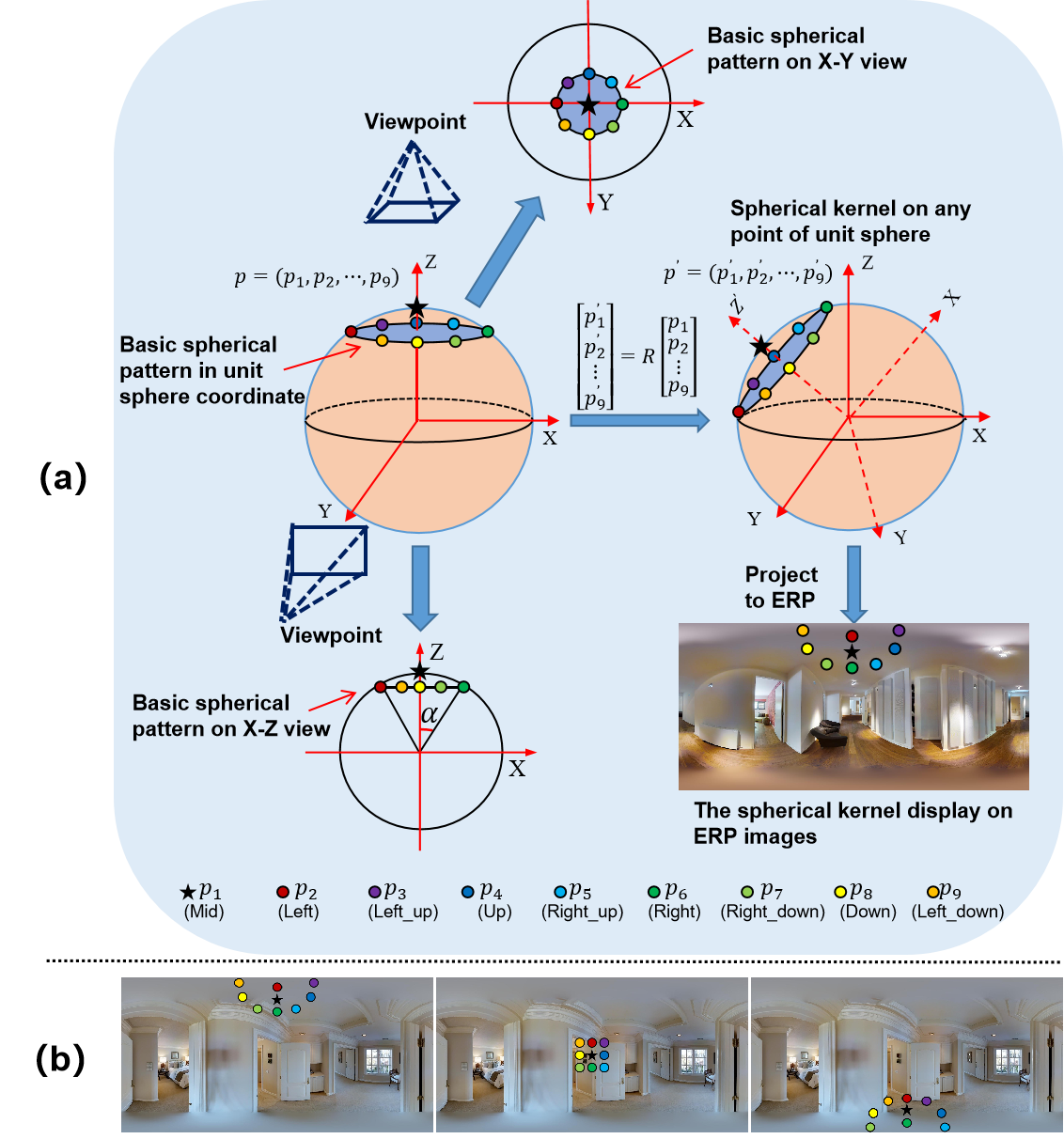}
  \caption{(a) Generation process of spherical convolution kernel. With a defined universal rotation matrix, spherical convolution kernels corresponding to different positions can be generated, which greatly reduces the computational cost. (b) Visualizing convolution kernels at the poles and equator positions in ERP images, which enable us to tackle distortion issues in distinct regions.}
  \label{Fig.3}
\end{figure}
A sphere is inherently a perfectly axis-symmetric shape, and it appears as a circle from any viewpoint(See \ref{Fig.3}(a)). 
The existing methods can be classified as three approaches: 1. Using a conventional square kernel for the generated plane tangent to the central point of a spherical model. 2.using the points of a discrete spherical image originating from a geodesic dome. 3. network training for the offset of sampling points. Different from them, our sampling is directly carried out on a sphere, which eliminates the problems of disconnectivity and distortion in contrast to an ERP image representation, results in a more natural circular kernel in contrast to a square kernel applied to a tangent plane, and a relatively more uniform sampling in contrast to a discrete spherical image originated from a geodesic dome. And more reliable compared to methods that depend on network predictions. Inspired by conventional feature point detection\cite{tracking}, we introduced a circular convolution kernels. In contrast to computing rectangular convolution kernels from tangent planes\cite{spherenet}\cite{OSRT}, circular kernels align more closely with the essence of a sphere and have the ability to extend beyond image boundaries. Any point on the sphere can be considered as the center of an infinite number of circles. Therefore, we choose a point on the sphere and the closest outer circle around it as the convolution kernel(See \ref{Fig.3}(a)).

When performing convolutions on a sphere, it is necessary to take into account the curvature and topological structure of the sphere, which increases the complexity of the convolution process. With an increase in latitude spacing, the impact will become more pronounced. However, by selecting the closest outer circle as the convolution kernel, it can preserve the geometric properties of the image and minimizing the boundary effects. A circle encompasses infinite points, it is difficult for practical calculations. Considering that planar convolutions typically employ 3x3 convolution kernels, we select eight equidistant points on the circle, along with the central point, as the spherical convolution kernel, as depicted in \ref{Fig.3}(a). In contrast to traditional discrete spherical sampling methods, which may sacrifice local detail to ensure global coverage, our method independently computes the convolution kernel for each point based on its adjacent points. This approach eliminates the requirement for a global discrete grid, leading to higher precision and making it more suitable for pixel-level prediction tasks.

Computing the coordinates$(x, y, z)$ of all pixels of an H×W ERP image projected onto a sphere, along with the corresponding coordinates on the outer circle, is a complex and time-consuming task. Therefore, we propose to define a basic spherical pattern, as illustrated in \ref{Fig.3}(a). Specifically, the outer circle is chosen as the basic spherical pattern at the North Pole of the sphere due to its unique geometric properties with coordinates (0, 0, 1).

An ERP image with H$\times$W is projected onto a unit sphere$(r=1)$, the distance between any two adjacent points on the equator is $\frac{2\pi}{W}$. Given the uniqueness of the sphere and the aspect ratio of the ERP image being 1:2, it follows that the distance between any two points on any circle centered at the sphere's center is also $\frac{2\pi}{W}$. As shown in \ref{Fig.3}, in the X${-}$Z view, let ${\alpha}$ denote the distance between any point on the circle and the Z-axis and ${\alpha}$ is $\frac{2\pi}{W}$.
The coordinates of the basic spherical pattern are as \ref{outer cycle} shows:
\begin{small}
\begin{equation}
\begin{aligned}
\label{outer cycle}
      p_1 &=(0,0,1)\\
      p_2,p_6 &=(0.sin(\alpha)r,\pm cos(\alpha)r)\\
      p_3,p_9&={(\pm sin(\frac{\pi}{4})sin(\alpha)r,cos(\frac{\pi}{4})sin(\alpha)r,cos(\alpha)r)}\\p_4,p_8 &=(\pm sin(\alpha)r,0,cos(\alpha)r) \\p_5,p_7&={(sin(\frac{\pi}{4})sin(\alpha)r,\pm cos(\frac{\pi}{4})sin(\alpha)r,cos(\alpha)r)}
\end{aligned}
\end{equation}
\end{small}
where $p_{1}$ denotes the North Pole point: $Mid$ , while $p_{2}$$\cdots$$p_{9}$ represents the points forming the base-spherical pattern $Left$, $Left_{up}$, $Up$, $Right_{up}$, $Right$, $Right_{down}$, $Down$ and $Left_{down}$ respectively.

As \ref{p} and \ref{Fig.3}(a) shows, by applying a same procedure of rotating the basic spherical pattern with a consistent rotation matrix $R$, we can effectively reposition the pattern on the sphere through spherical rotations. The employment of consistent rotation facilitates the generation of the convolution kernel at different positions, while ensuring that the distribution of points on the outer circle adheres to the original distribution of the basic spherical pattern in sphere. As shown in \ref{Fig.3}(b), the proposed spherical convolution kernel takes on different shapes in different regions of the image.
\begin{equation}
\label{p}    \left[\begin{matrix}p_1^\prime\\\begin{matrix}p_2^\prime\\\vdots\\\end{matrix}\\p_9^\prime\\\end{matrix}\right]=R\left[\begin{matrix}p_1\\\begin{matrix}p_2\\\vdots\\\end{matrix}\\p_9\\\end{matrix}\right]
\end{equation}
Where ${p_{1}}^\prime$, ${p_{2}}^\prime$$\cdots$${p_{9}}^\prime$represents the nine points that make up the spherical kernel: $Mid^\prime$, ${Left}^\prime$, ${Left_{up}}^\prime$, ${Up}^\prime$, ${Right_{up}}^\prime$, ${Right}^\prime$, ${Right_{down}}^\prime$, ${Down}^\prime$, ${Left_{down}}^\prime$. Note that the process does not induce any deformation to the spherical kernel.
\subsubsection{Rotated matrix computation}

It is imperative to ensure a consistent rotation pattern is used when rotating from the North Pole point to a given point on the sphere. Following \ref{thetaphi}, a point$(x, y, z)$on the sphere can be represented by ($\theta$, $\varphi$):
\begin{equation}
\theta=arccos\left(z\right),\varphi=arctan2(y,x)
\label{thetaphi}
\end{equation}
where $\theta\in[0,\pi]$ denotes the inclination between the positive half-axis of the Z-axis and a specific point, while $\varphi\in[0,2\pi)$ represents the azimuthal angle between the projection of the point on the X-Y plane and the positive half-axis of the X-axis. Based on $\theta$ and $\varphi$, we could infer the rotation matrix $R$ for each point.
The spherical convolution kernel at any point on the sphere can be obtained through \ref{p}.

\begin{figure}[t]
  \centering
  \includegraphics[width=0.95\linewidth]{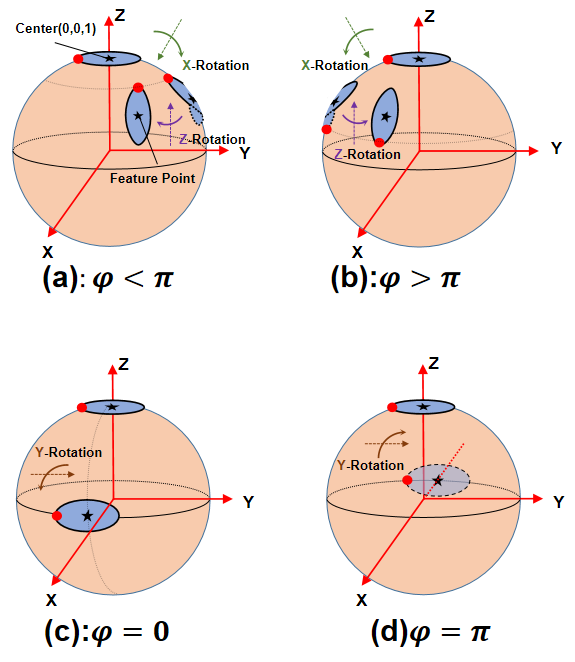}
  \caption{Rotation process}
  \label{r}
\end{figure}

We calculate the matrix $R$ for rotation around various axes based on the values of $\theta$ and $\varphi$. As depicted in \ref{juz}, $\varphi$ is partitioned into four distinct categories to accommodate different scenarios. As shown in \ref{r}(a),(b), when $\varphi$ is greater or less than $\pi$, the sphere rotates around fixed axes according to \ref{juz}. That is, it first rotates around the x-axis by $\theta$, then around the z-axis, and does not rotate around the y-axis. This allows the proposed pattern to be rotated to the desired position,  while maintaining the relative positions of points on the spherical pattern unchanged. As shown in \ref{r}(c),(d), when $\varphi$ equals $\pi$ or 0, the rotation solely occurs around the y-axis. This approach ensures that all points on the spherical surface rotate according to the same proposed rotation process.
\begin{equation}
\begin{aligned}
&\left\{\begin{matrix}yaw=\varphi-\frac{\pi}{2}\\pitch=0\\roll=-\theta\\\end{matrix},\varphi<\pi;\right.\left\{\begin{matrix}yaw=0\\pitch=-\theta\\roll=0\\\end{matrix},\varphi=\pi;\right.\\
&\left\{\begin{matrix}yaw=(\varphi-\pi)-\frac{\pi}{2}\\pitch=0\\roll=\theta\\\end{matrix},\varphi>\pi;\right.\left\{\begin{matrix}yaw=0\\pitch=\theta\\roll=0\\\end{matrix},\varphi=0;\right.
\label{juz}
\end{aligned}
\end{equation}
where $yaw$, $pitch$, and $roll$ denote the rotation angles around the Z-axis, Y-axis, and X-axis respectively. After obtaining the rotation angles, we can calculate the rotation matrices $R_X$, $R_Y$, $R_Z$ for each direction, and the final rotation matrix $R$ are as shown in \ref{r1}:
\begin{equation}
\begin{aligned}
\label{ypr}
    R_X&=\left[\begin{matrix}1&0&0\\0&cos(roll)&-sin(roll)\\0&sin(roll)&cos(roll)\\\end{matrix}\right];\\R_Y&=\left[\begin{matrix}cos(pitch)&0&sin(pitch)\\0&1&0\\-sin(piych)&0&cos(pitch)\\\end{matrix}\right];\\R_Z&=\left[\begin{matrix}cos(yaw)&-sin(yaw)&0\\sin(yaw)&cos(yaw)&0\\0&0&1\\\end{matrix}\right];  
\end{aligned}
\end{equation}
\begin{equation}
\label{r1}
    R=R_Z\bullet R_Y\bullet R_X
\end{equation}
Utilizing $R$ within the proposed basic spherical pattern enables the derivation of spherical convolution kernels that correspond to any position on the sphere.
\begin{figure}[h]
  \centering
  \includegraphics[width=0.98\linewidth]{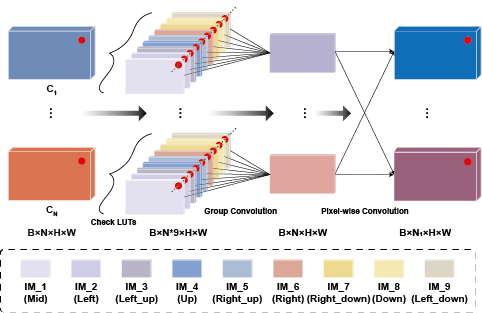}
\caption{The relative position of the spherical convolution kernel for each pixel in the image is stored in the corresponding LUTs, which in turn maps the ERP image to nine sub-images. Then, after group convolution and pixel-wise convolution, an $R^{N_{1}\times H\times W}$ feature map is obtained.}
  \label{Fig.4}
\end{figure}
\subsubsection{Separable spherical kernel convolution}
For convolution, the network assigns unique weights to each channel's convolution kernel and accomplishes the convolution by moving these kernels over the image. However, on the sphere, convolution kernels are not translated but instead rotated. Previous methods employed grids to implement spherical convolution. While the grid-based approaches are constrained by the number of grids used, and may fail to achieve per-pixel division, which is detrimental for pixel-level tasks. To address this issue, we introduce a per-pixel separable spherical kernel convolution method. As shown in \ref{Fig.4}, We firstly maps spherical convolution kernels, centered at each pixel on the image, to the same position in different images. Subsequently, we conduct group convolution with a size of 1, where all pixels comprising the kernel in the convolution process are grouped together. This operation eliminates the need for additional padding to understand image boundaries. For the pixel-wise task, we believe that the introduced pixel-wise convolution strengthens the sensitivity for our network to inter dependencies among neighboring pixels on the sphere, enhancing the capacity of the network to perceive structural information in panoramic images.

specificly, we propose to use look-up tables(LUTs) to store the respective relative positions of the spherical convolution kernel at each point. For instance, LUT1 stores the positions of $p_1$ (i.e., the '$Mid$' point) for each pixel of the image, which is the center of the proposed spherical convolution kernel, and LUT2 stores the relative positions of $p_2$ (i.e., the '$Left$' point) in the spherical convolution kernel for each pixel of the image. With LUT2, we can obtain an image that is entirely composed of $p_2$ from a given image while maintaining the size of the original image. Similarly, LUT3 to LUT9 represent the positions of $p_3$ to $p_9$ in the spherical convolution kernel (see \ref{Fig.4}).

As depicted in \ref{Fig.4}, once the LUTs are available, a feature$ {R^{N\times H\times W}}$ can be mapped to ${R^{N*9\times H\times W}}$ sub-features from IM\_1 to IM\_9. Based on it, the group convolution with a kernel size of 1 can be employed to equivalently replace the original kernel convolution. Finally, a pixel-wise convolution is conducted to expand the channels in the spherical convolution, and a feature$\in R^{N_1\times H\times W}$(where $N_1$ can be arbitrarily set) can be obtained. 

\subsection{Segmentation feature fusion module}
The feature $f_{ERP}$ obtained through planar convolution exhibits distortion at the poles, while the feature $f_{sph}$ obtained through the proposed spherical convolution method is distortion-free. The effectiveness and importance of planar convolution has been demonstrated in this task\cite{14}, research on the reliability of spherical convolution in deeper layers remains limited. To exploit the advantages of both convolutions, we propose to integrate $f_{sph}$ into $f_{ERP}$. Given the minimal curvature near the equator in panoramic images, the distortion in this region is negligible. Therefore, planar convolution in this area is reasonable. In the panoramic image domain, researchers typically assume significant distortion in the upper and lower thirds, with almost no distortion in the central part. Leveraging the well-established method of planar convolution enables effective feature extraction from panoramic images, we preserve the features extracted through planar convolution near the equator. This strategy enables us to fully harness the benefits of planar convolution in extracting rich features from the image while avoiding the potential adverse impact of spherical convolution in deep layers.
\begin{figure*}[t]
  \centering
  \includegraphics[width=0.95\linewidth]{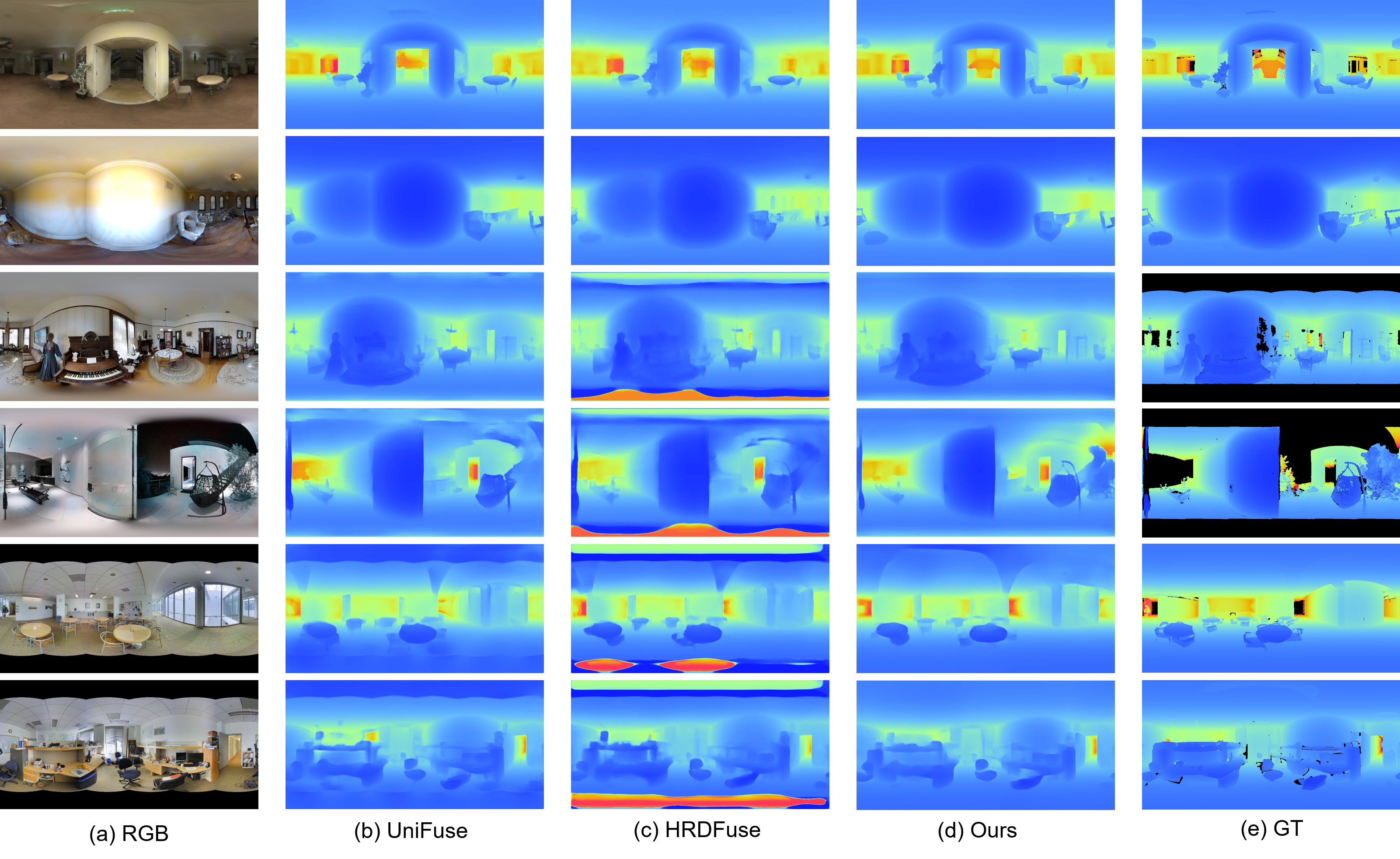}
  \caption{Results of qualitative comparison on 3D60 (top), Matterport3D (middle) and Stanford2D3D (bottom).}
  \label{Fig.5}
\end{figure*}
As shown in \ref{Fig.2}, we segment $f_{ERP}$ into three equal parts and discard the features at the North and South poles, retaining the features in the middle part ($f_{ERP-mid}$). To fuse $f_{ERP}$ and $f_{sph}$, we propose an adaptive weight fusion scheme, where we perform adaptive fusion on the two features to obtain an initial fused feature $f$:
\begin{equation}
    f=w_0\times f_{ERP}+w_1\times f_{sph}
\end{equation}
where $w_0$ and $w_1$ are learnable parameters. Then, we prioritize $f_{ERP}$ as the primary carrier and perform a concatenation operation between $f$ and $f_{ERP}$. Subsequently, a convolution layer is used to achieve fusion and retain the feature $f_{ERP-mid}$ extracted near the equator. Lastly, a simple non-linear activation is applied to obtain the final fused feature $f_{fused}$.
The fused feature $f_{fused}$ effectively combines the superior features extracted by $f_{ERP-mid}$ near the equator with the distortion-free features $f_{sph}$.

\subsection{teacher network}
The proposed teacher-student network, as depicted in \ref{Fig.2}, aims to incorporate more depth information into the network by utilizing ground truth depth and compensating for the shortcomings of spherical convolution in deep layers.

The teacher network takes the ground truth depth as input, generating the latent features in the deepest layer, which acts as guidance for the student model. By leveraging the inherent characteristics of the teacher model, we can enrich the depth information contained in the latent features of the student model, thereby improving the network's performance in depth estimation. It is important to note that the teacher network is discarded during the final inference. During the training of the teacher model, we employ the commonly used Burhu loss\cite{laina2016deeper} as the loss function for depth estimation tasks.
\section{Experiments}
\label{sec:experiment}
\begin{table*}[t]
  \centering
    \resizebox{0.90\textwidth}{!}{
     \begin{tabular}{c|c|c|c|c|c|c|c|c}
        \toprule
        Datasets & Method & Abs Rel$\downarrow$ & Sq Rel$\downarrow$ & RMSE$\downarrow$ & RMSE(log)$\downarrow$ & $\delta_{1}\uparrow$ & $\delta_{2}\uparrow$ & $\delta_{3}\uparrow$\\
        \midrule
        \multirow{9}{*}{Standford2D3D}   &  FCRN\cite{laina2016deeper} &  \hspace{1.3em}-\hspace{1.3em}/ 0.1837     &    \hspace{1.3em}-\hspace{1.3em}/\hspace{1.3em}-\hspace{1.3em}    &     \hspace{1.3em}-\hspace{1.3em}/ 0.5774    &    \hspace{1.3em}-\hspace{1.3em}/\hspace{1.3em}-\hspace{1.3em}    &    \hspace{1.3em}-\hspace{1.3em}/ 0.7230     &    \hspace{1.3em}-\hspace{1.3em}/ 0.9207    &    \hspace{1.3em}-\hspace{1.3em}/ 0.9731\\
         & BiFuse with fusion\cite{13}     	& \hspace{1.3em}-\hspace{1.3em}/ 0.1209    &    \hspace{1.3em}-\hspace{1.3em}/\hspace{1.3em}-\hspace{1.3em}    &     \hspace{1.3em}-\hspace{1.3em}/ 0.4142    &    \hspace{1.3em}-\hspace{1.3em}/\hspace{1.3em}-\hspace{1.3em}    &    \hspace{1.3em}-\hspace{1.3em}/ 0.8660    &    \hspace{1.3em}-\hspace{1.3em}/ 0.9580    &   \hspace{1.3em}-\hspace{1.3em}/ 0.9860 \\
         & UniFuse with fusion\cite{14}    	& \hspace{1.3em}-\hspace{1.3em}/ 0.1114    &    \hspace{1.3em}-\hspace{1.3em}/\hspace{1.3em}-\hspace{1.3em}    &     \hspace{1.3em}-\hspace{1.3em}/ 0.3691    &    \hspace{1.3em}-\hspace{1.3em}/ 0.2322    &    \hspace{1.3em}-\hspace{1.3em}/ 0.8711    &    \hspace{1.3em}-\hspace{1.3em}/ 0.9664    &    \hspace{1.3em}-\hspace{1.3em}/ 0.9882\\
         & OmniFusion (2-iter)\cite{15}   	& 0.0950 /\hspace{1.3em}-\hspace{1.3em}    & 0.0491 /\hspace{1.3em}-\hspace{1.3em}    &    0.3474 /\hspace{1.3em}-\hspace{1.3em}  &    0.1599 /\hspace{1.3em}-\hspace{1.3em}    &    0.8988 /\hspace{1.3em}-\hspace{1.3em}    &    0.9769 /\hspace{1.3em}-\hspace{1.3em}    & 0.9924 /\hspace{1.3em}-\hspace{1.3em} \\
         & PanoFormer*\cite{panoformer} 		& \hspace{1.3em}-\hspace{1.3em}/ 0.1131  &  \hspace{1.3em}-\hspace{1.3em}/ 0.0723   & \hspace{1.3em}-\hspace{1.3em}/ 0.3557  & \hspace{1.3em}-\hspace{1.3em}/ 0.2454   &  \hspace{1.3em}-\hspace{1.3em}/ 0.8808  &  \hspace{1.3em}-\hspace{1.3em}/ 0.9623   & \hspace{1.3em}-\hspace{1.3em}/ 0.9855  \\ 
         & SphereDepth\cite{17} 	& \hspace{1.3em}-\hspace{1.3em}/\hspace{1.3em}-\hspace{1.3em}  & \hspace{1.3em}-\hspace{1.3em}/\hspace{1.3em}-\hspace{1.3em}  &  \hspace{1.3em}-\hspace{1.3em}/ 0.4512 &  \hspace{1.3em}-\hspace{1.3em}/\hspace{1.3em}-\hspace{1.3em} & \hspace{1.3em}-\hspace{1.3em}/ 0.8666 &  \hspace{1.3em}-\hspace{1.3em}/ 0.9642 &  \hspace{1.3em}-\hspace{1.3em}/ 0.9863 \\
         & PanelNet\cite{PanelNet} 		& \hspace{1.3em}-\hspace{1.3em} /\hspace{1.3em}-\hspace{1.3em} & \hspace{1.3em}-\hspace{1.3em} /\hspace{1.3em}-\hspace{1.3em}  & 0.2933 /\hspace{1.3em}-\hspace{1.3em} & \hspace{1.3em}-\hspace{1.3em} /\hspace{1.3em}-\hspace{1.3em} & 0.9242 /\hspace{1.3em}-\hspace{1.3em} & 0.9796 /\hspace{1.3em}-\hspace{1.3em} & 0.9915 /\hspace{1.3em}-\hspace{1.3em} \\
         & HRDFuse\cite{16} 		& 0.0935 /\hspace{1.3em}-\hspace{1.3em} & 0.0508 /\hspace{1.3em}-\hspace{1.3em}  & 0.3106 /\hspace{1.3em}-\hspace{1.3em} & 0.1422 /\hspace{1.3em}-\hspace{1.3em} & 0.9140 /\hspace{1.3em}-\hspace{1.3em} & 0.9798 /\hspace{1.3em}-\hspace{1.3em} & 0.9927 /\hspace{1.3em}-\hspace{1.3em} \\
        \midrule
         & Ours & \textbf{{0.0926}} / \textbf{{0.0940}} & \textbf{0.0487} / \textbf{{0.0541}} & 0.3058 / \textbf{{0.3269}} & \textbf{{0.1396}} / \textbf{{0.1417}} & 0.9188 / \textbf{{0.9143}} & \textbf{{0.9804}} / \textbf{{0.9808}} & \textbf{{0.9931}} / \textbf{{0.9921}} \\
        \cmidrule{2-9}
         &  Teacher Network & 0.0086 / 0.0093 & 0.0013 / 0.0021 &  0.0608 / 0.0758 & 0.0214 / 0.0270 &  0.9983 / 0.9971 &  0.9997 / 0.9994 &  0.9999 / 0.9998\\  
        \bottomrule
        \bottomrule
        
        \multirow{9}{*}{3D60}   &  FCRN\cite{laina2016deeper} &  \hspace{1.3em}-\hspace{1.3em}/ 0.0699     &    \hspace{1.3em}-\hspace{1.3em}/ 0.2833     &    \hspace{1.3em}-\hspace{1.3em}/\hspace{1.3em}-\hspace{1.3em}    &      \hspace{1.3em}-\hspace{1.3em}/\hspace{1.3em}-\hspace{1.3em}     &      \hspace{1.3em}-\hspace{1.3em}/ 0.9532     &    \hspace{1.3em}-\hspace{1.3em}/ 0.9905    &    \hspace{1.3em}-\hspace{1.3em}/ 0.9966\\
         & BiFuse with fusionp\cite{13}     	& \hspace{1.3em}-\hspace{1.3em}/ 0.0615    &    \hspace{1.3em}-\hspace{1.3em}/\hspace{1.3em}-\hspace{1.3em}    &    \hspace{1.3em}-\hspace{1.3em}/ 0.2440    &    \hspace{1.3em}-\hspace{1.3em}/\hspace{1.3em}-\hspace{1.3em}    &    \hspace{1.3em}-\hspace{1.3em}/ 0.9699    &    \hspace{1.3em}-\hspace{1.3em}/ 0.9927  &     \hspace{1.3em}-\hspace{1.3em}/ 0.9969 \\
         & UniFuse with fusion\cite{14}   & 	\hspace{1.3em}-\hspace{1.3em}/ 0.0466    &    \hspace{1.3em}-\hspace{1.3em}/\hspace{1.3em}-\hspace{1.3em}    &     \hspace{1.3em}-\hspace{1.3em}/ 0.1968    &    \hspace{1.3em}-\hspace{1.3em}/ 0.0725    &    \hspace{1.3em}-\hspace{1.3em}/ 0.9835    &   \hspace{1.3em}-\hspace{1.3em}/ 0.9965    &    \hspace{1.3em}-\hspace{1.3em}/ 0.9987\\
         & OmniFusion (2-iter)\cite{15}   	& 0.0430 /\hspace{1.3em}-\hspace{1.3em}    &       0.0114 /\hspace{1.3em}-\hspace{1.3em}  &  0.1808 /\hspace{1.3em}-\hspace{1.3em}   &    0.0735 /\hspace{1.3em}-\hspace{1.3em}    &    0.9859 /\hspace{1.3em}-\hspace{1.3em}     &     0.9969 /\hspace{1.3em}-\hspace{1.3em}    &    0.9989 /\hspace{1.3em}-\hspace{1.3em} \\
         & ODE-CNN\cite{26} 		&  \hspace{1.3em}-\hspace{1.3em}/ 0.0467   &    \hspace{1.3em}-\hspace{1.3em}/ 0.0124  & \hspace{1.3em}-\hspace{1.3em}/ 0.1728  & \hspace{1.3em}-\hspace{1.3em}/ 0.0793  &  \hspace{1.3em}-\hspace{1.3em}/ 0.9814 & \hspace{1.3em}-\hspace{1.3em}/ 0.9967  & \hspace{1.3em}-\hspace{1.3em}/ 0.9889 \\ 
         &  SphereDepth\cite{17} &	\hspace{1.3em}-\hspace{1.3em}/ 0.0550    &    \hspace{1.3em}-\hspace{1.3em}/ 0.1145  &  \hspace{1.3em}-\hspace{1.3em}/ 0.2364 &  \hspace{1.3em}-\hspace{1.3em}/\hspace{1.3em}-\hspace{1.3em} &  \hspace{1.3em}-\hspace{1.3em}/ 0.9743 &  \hspace{1.3em}-\hspace{1.3em}/ 0.9944   &  \hspace{1.3em}-\hspace{1.3em}/ 0.9978\\
         & HRDFuse\cite{16} 		& 0.0358 /\hspace{1.3em}-\hspace{1.3em}   &    0.0100 /\hspace{1.3em}-\hspace{1.3em}  & 0.1555 /\hspace{1.3em}-\hspace{1.3em} & 0.0592 /\hspace{1.3em}-\hspace{1.3em} & 0.9894 /\hspace{1.3em}-\hspace{1.3em} & 0.9973 /\hspace{1.3em}-\hspace{1.3em} & 0.9990 /\hspace{1.3em}-\hspace{1.3em} \\
        \midrule
         & Ours & 0.0394 / \textbf{{0.0379}} & 0.0101 / \textbf{{0.0105}} & 0.1560 / \textbf{{0.1687}} & 0.0604 / \textbf{{0.0602}} & \textbf{{0.9897}} / \textbf{{0.9901}} & \textbf{{0.9975}} / \textbf{{0.9975}} & \textbf{{0.9990}} / \textbf{{0.9991}} \\  
         \cmidrule{2-9}
          &  Teacher Network & 0.0081 / 0.0051 & 0.0005 / 0.0004 &   0.0401 / 0.0380  &   0.0143 /  0.0054 & 0.9996 /   0.9996 &   0.9999 /   0.9999 &   0.9999 / 0.9999\\  
        \bottomrule
        \bottomrule
        \multirow{9}{*}{Matterport3D}   &  FCRN\cite{laina2016deeper} &  0.2409     &    -     &     0.6704    &    -    &    0.7703    &   0.9174    &    0.9617\\
      
                & BiFuse with fusion\cite{13}     	& 0.2048   &    -   &     0.6259    &    -  &   0.8452    &    0.9319   &    0.9632\\
       
               & UniFuse with fusion\cite{14}   	& 0.1063    &    -   &     0.4941    &    0.1613    &    0.8897    &    0.9623   &    0.9831\\
    	
               & OmniFusion (2-iter)*\cite{15}   	& 0.1007    &   0.0969   &    0.4435  &   0.1664    &    0.9143   &    0.9666   &   0.9844\\
         
              &  PanoFormer*\cite{panoformer} 		& 0.0904    &   0.0764   &    0.4470  &   0.1650    &    0.8816   &    0.9661   &   0.9878\\ 
      
               & SphereDepth\cite{17} 	&    -    &    -    &    0.5922  &   -    &    0.8620   &    0.9519   &   0.9770\\
               & PanelNet\cite{PanelNet} 		& -    &  -   &    0.4528  &   -    &    0.9123   &    0.9703   &   0.9856\\
               & HRDFuse\cite{16} 		& 0.0967    &   0.0936   &    0.4433  &   0.1642    &    0.9162   &    0.9669   &   0.9844\\

              \midrule
      
             &  Ours & 0.0941 & \textbf{{0.0723}}  & \textbf{{0.4396}}  & \textbf{0.1402} & 0.9110 & \textbf{{0.9712}} & \textbf{{0.9904}} \\
             \cmidrule{2-8}\cmidrule{9-9}
              &  Teacher Network & 0.0186 & 0.0049  &   0.1262 &   0.0162 &    0.9954 &   0.9991 &   0.9997\\ 
             \bottomrule
      \end{tabular}}
\caption{Quantitative comparison with other methods. Bold indicates that our method performs the best. -/-: On the left side of /, it indicates that the dataset processing estimates a depth of 8 meters, while on the right side of /, it indicates that the dataset processing estimates a depth of 10 meters. *:It indicates that due to the absence of a pre-trained model, its metrics are derived from the latest SOTA model, \cite{16}.}
\label{Table.1}
\end{table*}
\subsection{Datesets, Metrics and Implimentation details}
\textbf{Datesets:} In this paper, We conducted experiments on three benchmark datasets that are widely used for this tasks: 3D60\cite{3d60}, Matterport3D\cite{m3d}, and Stanford2D3D\cite{s2d3d} datasets. Stanford2D3D and Matterport3Dare real-world datasets. While 3D60\cite{3d60} is composed of two synthetic datasets: SUNCG\cite{sun} and SceneNet\cite{sce} and two real-world datasets: Stanford2D3D and Matterport3D. Note that there are some rendering issues\cite{14} with the 3D60, and some anomalies may occur in this task.\\
\textbf{Metrics:} Following previous work\cite{14,16}, we adopt standard evaluation metrics for evaluation: Absolute Relative Error (Abs Rel), Squared Relative Error (Sq Rel), Root Mean Squared Error (RMSE), Root Mean Squared Error in logarithmic space (RMSE(log)) and accuracy with a threshold $\delta_{t}$, where $t$ $\in$ \{$1.25$,${1.25}^2$,${1.25}^3$\}.\\
\textbf{Implimentation details:} Our network was trained using the Adam optimizer, a batch size of 1, and a learning rate of $1\times10^{-4}$ on a TITAN RTX 24G. 
We trained our model for only 30 epochs for Matterport3D, 3D60 and 20 epochs for 
Stanford2D3D. Moreover, we adopt augmentation techniques, random color adjustment, and left-right-flipping, random yaw rotation in the training phase.
\subsection{comparision with state of the art}
\ref{Table.1} presents a comparative analysis between our method and existing methods for depth estimation. Notably, Some methods like \cite{16} and \cite{15} differ from conventional depth estimation methods in terms of data processing for the Stanford2D3D and 3D60 datasets. Specifically, the training data and testing data have a maximum depth of 8 meters for these two datasets, while traditional methods like \cite{14} \cite{laina2016deeper} and \cite{17} have a maximum depth of 10 meters. In order to analyze the results more comprehensively and to adequately compare our method with other methods, we evaluated the two different depth estimation results(8m and 10m) for our method. For the Matterport3D dataset, all existing methods have the same maximum depth value of 10. Here we clarify that due to the unavailability of pre-trained models for some methods (e.g., Omnifuse does not provide a pre-trained model for the Matterport3D dataset, and Panoformer, PanelNet and HRDFuse does not provide any pre-trained models), for fair comparisons, we collected publicly available experimental data of competitors from the comparisons made by the latest SOTA depth estimation model HRDfuse.

As \ref{Table.1} shows, Our method performs well compared to SOTA methods\cite{13,14,15,16,17,panoformer,PanelNet} on several benchmark datasets. On the Stanford2D3D dataset, our method outperforms Unifuse by 17.5\% (Abs Rel) and 12.91\% (RMSE), outperforms Omnifuse by 2.59\%(Abs Rel) and 13.6\%(RMSE),our method outperforms Panlenet by 0.082\% ($\delta_1$) and 0.16\% ($\delta_3$), and outperforms HRDfuse by 0.972\%(abs rel), 1.57\% (RMSE) and 0.525\% ($\delta_1$). On the 3D60 dataset, our method outperforms Unifuse by 22.96\% (abs rel) and 16.66\% (RMSE), outperforms Omnifuse by 9.14\% (abs rel), outperforms ODE-CNN by 23.22\%(Abs Rel) and 2.43\%(RMSE), and outperforms HRDfuse with 0.03\% ($\delta_1$), while also demonstrating competitive results on other metrics with HRDFuse. Furthermore, it is observed that the method introduced in this paper achieves a slightly superior of accuracy in comparison to HRDFuse. On the Matterport3D dataset, our method outperforms Unifuse by 18.38\% (Abs Rel) and 12.37\% (RMSE), outperforms Omnifuse by 7\% (Abs Rel), outperforms PanelNet by 0.485\% ($\delta_3$)  and over 3\% (RMSE), and outperforms HRDfuse by 2.76\% (Abs Rel), 29.46\% (Sq Rel), 0.841\% (RMSE) and 0.61\%($\delta_3$). In \ref{Fig.5}, since HRDFuse does not provide any pre-trained models, we retrained the model to the official Settings for visualization, and we qualitatively compare our method with UniFuse and HRDFuse, and our method outperforms them. 
\subsection{ablation study}
\subsubsection{ablation study of each component}
We conducted a series of incremental experiments to assess the effectiveness of each component, as illustrated in \ref{Table.2}. The ablation experiment was performed with maximum depth of 8 meters on the Stanford2D3D. We used only the planar convolution method for depth estimation as the baseline. Subsequently, we added the proposed spherical convolution method, teacher network, and SFF module sequentially.
\begin{table}[t]
  \centering
  \resizebox{0.9\linewidth}{!}{
  \begin{tabular}{c|c|c|c|c|c|c|c}
    \toprule
        Base & S-Conv & Teacher & SFF & Abs Rel$\downarrow$ & Sq Rel$\downarrow$ & RMSE$\downarrow$ & $\delta_{1}\uparrow$\\
         \midrule
          $\checkmark$ & & & & 0.1125 & 0.0599 & 0.3434 & 0.8870 \\     
          \midrule
          $\checkmark$ & $\checkmark$ & & &  0.1050 & 0.0564 & 0.3239 & 0.9066\\
          
          \midrule
          $\checkmark$ & $\checkmark$ & $\checkmark$ & & 0.0968 & 0.0546 & 0.3124 & 0.9084\\
          
         \midrule
         $\checkmark$ & $\checkmark$ & & $\checkmark$ & 0.0986 & 0.0507 & 0.3131 & 0.9156\\
         
         \midrule
        $\checkmark$ & $\checkmark$ & $\checkmark$ & $\checkmark$ & \textbf{0.0926} & \textbf{0.0487} & \textbf{0.3058} & \textbf{0.9188}\\
        
    \bottomrule
   \end{tabular}
  }
  \caption{Ablation study for different combinations of independent components.}
  \label{Table.2}
\end{table}
As shown in \ref{Table.2}, the performance of the planar convolution model was adversely affected by distortion. With the introducing of proposed the spherical convolution method, the performance improved by 6.21\% (Sq Rel). However, we only used a simple concatenation method for fusion, which significantly reduced fusion effectiveness, while the performance has greatly improved by 11.24\% (Sq Rel) since we utilized the SFF module. Moreover, we assessed the effectiveness of the teacher network by incorporating it into the network without the SFF module, resulting in a 3.3\% (Sq Rel) improvement. Finally, when all components were used, the performance achieved the maximum improvement of 23.00\%(Sq Rel). The experimental results illustrate that each proposed component plays a pivotal role in this task, notably elevating the overall performance of the network.
\subsubsection{weight of fusion}
We performed ablation experiments on the weights of SFF module, as presented in \ref{Table.3}. We assigned fixed weight ratios of 1:0, 0:1, and 0.5:0.5, in addition to using adaptive weights. The experimental results demonstrate that the adaptive weights outperform the other three fixed weight ratios. Overall, the results provide further evidence of the effectiveness and reliability of the proposed SFF module.
\begin{table}[t]
  \centering
   \resizebox{0.9\linewidth}{!}{
    \begin{tabular}{c|c|c|c|c|c}
    \toprule
      ERP feature & Spherical feature & Abs Rel$\downarrow$ & Sq Rel$\downarrow$ & RMSE$\downarrow$  & $\delta_{3}\uparrow$\\
      \midrule
      0.5 & 0.5 &     0.1045   &   0.0535   &   0.3112   &   0.9930 \\
      
      0   & 1   &     0.0928   &   0.0510   &   0.3059   &   0.9927 \\
    
      1   & 0   &     0.1011   &   0.0533   &   0.3139   &  0.9929\\
      \midrule
      \multicolumn{2}{c|}{Adaptive weighting}  & \textbf{0.0926} & \textbf{0.0487} & \textbf{0.3058} & \textbf{0.9931} \\
    \bottomrule
  \end{tabular}
    }
  \caption{The Ablation study on the weight of SFF module.}
  \label{Table.3}
\end{table}
\section{Conclusions and future work}
\label{sec:conclusion}
In this paper, we propose a method of depth estimation of a monocular panoramic image.
To the best of our knowledge, it is the first of fusing equirectangular and spherical representations so as to mitigate the effect of the disconnectivity and distortion of ERP images, and supervise the student network to learn the inherent cues of depth images of training set via a teacher-student model.
The experiments  shows the effectiveness of the proposed method.
Since depth estimation is a basic technique for image understanding, we believe the proposed method can find a lot of applications, such as visual surveillance, robot navigation and so on.
It is also our future work to do.


\begin{thebibliography}{10}

\bibitem{22}
Hong-Xiang Chen, Kunhong Li, Zhiheng Fu, Mengyi Liu, Zonghao Chen, and Yulan Guo.
\newblock Distortion-aware monocular depth estimation for omnidirectional images.
\newblock {\em IEEE Signal Processing Letters}, 28:334--338, 2021.

\bibitem{23}
Keisuke Tateno, Nassir Navab, and Federico Tombari.
\newblock Distortion-aware convolutional filters for dense prediction in panoramic images.
\newblock In {\em Proceedings of the European Conference on Computer Vision (ECCV)}, pages 707--722, 2018.

\bibitem{acdnet}
Chuanqing Zhuang, Zhengda Lu, Yiqun Wang, Jun Xiao, and Ying Wang.
\newblock Acdnet: Adaptively combined dilated convolution for monocular panorama depth estimation.
\newblock In {\em Proceedings of the AAAI Conference on Artificial Intelligence}, volume~36, pages 3653--3661, 2022.

\bibitem{slice}
Giovanni Pintore, Marco Agus, Eva Almansa, Jens Schneider, and Enrico Gobbetti.
\newblock Slicenet: deep dense depth estimation from a single indoor panorama using a slice-based representation.
\newblock In {\em Proceedings of the IEEE/CVF Conference on Computer Vision and Pattern Recognition}, pages 11536--11545, 2021.

\bibitem{17}
Qingsong Yan, Qiang Wang, Kaiyong Zhao, Bo~Li, Xiaoweo Chu, and Fei Deng.
\newblock Spheredepth: Panorama depth estimation from spherical domain.
\newblock In {\em 2022 International Conference on 3D Vision (3DV)}, pages 1--10. IEEE, 2022.

\bibitem{20}
Nikolaos Zioulis, Antonis Karakottas, Dimitrios Zarpalas, Federico Alvarez, and Petros Daras.
\newblock Spherical view synthesis for self-supervised 360 depth estimation.
\newblock In {\em 2019 International Conference on 3D Vision (3DV)}, pages 690--699. IEEE, 2019.

\bibitem{21}
Yuyan Li, Zhixin Yan, Ye~Duan, and Liu Ren.
\newblock Panodepth: A two-stage approach for monocular omnidirectional depth estimation.
\newblock In {\em 2021 International Conference on 3D Vision (3DV)}, pages 648--658. IEEE, 2021.

\bibitem{11}
Yu-Chuan Su and Kristen Grauman.
\newblock Learning spherical convolution for fast features from 360 imagery.
\newblock {\em Advances in Neural Information Processing Systems}, 30, 2017.

\bibitem{12}
Shang-Ta Yang, Fu-En Wang, Chi-Han Peng, Peter Wonka, Min Sun, and Hung-Kuo Chu.
\newblock Dula-net: A dual-projection network for estimating room layouts from a single rgb panorama.
\newblock In {\em Proceedings of the IEEE/CVF Conference on Computer Vision and Pattern Recognition}, pages 3363--3372, 2019.

\bibitem{10}
Hsien-Tzu Cheng, Chun-Hung Chao, Jin-Dong Dong, Hao-Kai Wen, Tyng-Luh Liu, and Min Sun.
\newblock Cube padding for weakly-supervised saliency prediction in 360 videos.
\newblock In {\em Proceedings of the IEEE Conference on Computer Vision and Pattern Recognition}, pages 1420--1429, 2018.

\bibitem{13}
Fu-En Wang, Yu-Hsuan Yeh, Min Sun, Wei-Chen Chiu, and Yi-Hsuan Tsai.
\newblock Bifuse: Monocular 360 depth estimation via bi-projection fusion.
\newblock In {\em Proceedings of the IEEE/CVF Conference on Computer Vision and Pattern Recognition}, pages 462--471, 2020.

\bibitem{14}
Hualie Jiang, Zhe Sheng, Siyu Zhu, Zilong Dong, and Rui Huang.
\newblock Unifuse: Unidirectional fusion for 360 panorama depth estimation.
\newblock {\em IEEE Robotics and Automation Letters}, 6(2):1519--1526, 2021.

\bibitem{9}
Marc Eder, Mykhailo Shvets, John Lim, and Jan-Michael Frahm.
\newblock Tangent images for mitigating spherical distortion.
\newblock In {\em Proceedings of the IEEE/CVF Conference on Computer Vision and Pattern Recognition}, pages 12426--12434, 2020.

\bibitem{16}
Hao Ai, Zidong Cao, Yan-Pei Cao, Ying Shan, and Lin Wang.
\newblock Hrdfuse: Monocular 360deg depth estimation by collaboratively learning holistic-with-regional depth distributions.
\newblock In {\em Proceedings of the IEEE/CVF Conference on Computer Vision and Pattern Recognition}, pages 13273--13282, 2023.

\bibitem{3d60}
Nikolaos Zioulis, Antonis Karakottas, Dimitrios Zarpalas, and Petros Daras.
\newblock Omnidepth: Dense depth estimation for indoors spherical panoramas.
\newblock In {\em Proceedings of the European Conference on Computer Vision (ECCV)}, pages 448--465, 2018.

\bibitem{m3d}
Angel Chang, Angela Dai, Thomas Funkhouser, Maciej Halber, Matthias Niessner, Manolis Savva, Shuran Song, Andy Zeng, and Yinda Zhang.
\newblock Matterport3d: Learning from rgb-d data in indoor environments.
\newblock {\em arXiv preprint arXiv:1709.06158}, 2017.

\bibitem{s2d3d}
Iro Armeni, Sasha Sax, Amir~R Zamir, and Silvio Savarese.
\newblock Joint 2d-3d-semantic data for indoor scene understanding.
\newblock {\em arXiv preprint arXiv:1702.01105}, 2017.

\bibitem{24}
Jiayang Bai, Shuichang Lai, Haoyu Qin, Jie Guo, and Yanwen Guo.
\newblock Glpanodepth: Global-to-local panoramic depth estimation.
\newblock {\em arXiv preprint arXiv:2202.02796}, 2022.

\bibitem{15}
Yuyan Li, Yuliang Guo, Zhixin Yan, Xinyu Huang, Ye~Duan, and Liu Ren.
\newblock Omnifusion: 360 monocular depth estimation via geometry-aware fusion.
\newblock In {\em Proceedings of the IEEE/CVF Conference on Computer Vision and Pattern Recognition}, pages 2801--2810, 2022.

\bibitem{PanelNet}
Haozheng Yu, Lu~He, Bing Jian, Weiwei Feng, and Shan Liu.
\newblock Panelnet: Understanding 360 indoor environment via panel representation.
\newblock In {\em 2023 IEEE/CVF Conference on Computer Vision and Pattern Recognition (CVPR)}, pages 878--887, 2023.

\bibitem{panoformer}
Zhijie Shen, Chunyu Lin, Kang Liao, Lang Nie, Zishuo Zheng, and Yao Zhao.
\newblock Panoformer: Panorama transformer for indoor 360 depth estimation.
\newblock In {\em European Conference on Computer Vision}, pages 195--211. Springer, 2022.

\bibitem{EGformer}
Ilwi Yun, Chanyong Shin, Hyunku Lee, Hyuk-Jae Lee, and Chae~Eun Rhee.
\newblock Egformer: Equirectangular geometry-biased transformer for 360 depth estimation.
\newblock In {\em 2023 IEEE/CVF International Conference on Computer Vision (ICCV)}, pages 6078--6089, 2023.

\bibitem{26}
Ming Li, Xuejiao Hu, Jingzhao Dai, Yang Li, and Sidan Du.
\newblock Omnidirectional stereo depth estimation based on spherical deep network.
\newblock {\em Image and Vision Computing}, 114:104264, 2021.

\bibitem{28}
Jie Li, Ling Han, Chong Zhang, Qiyue Li, and Zhi Liu.
\newblock Spherical convolution empowered viewport prediction in 360 video multicast with limited fov feedback.
\newblock {\em ACM Transactions on Multimedia Computing, Communications and Applications}, 19(1):1--23, 2023.

\bibitem{29}
Chenglei Wu, Ruixiao Zhang, Zhi Wang, and Lifeng Sun.
\newblock A spherical convolution approach for learning long term viewport prediction in 360 immersive video.
\newblock In {\em Proceedings of the AAAI Conference on Artificial Intelligence}, volume~34, pages 14003--14040, 2020.

\bibitem{OSRT}
Fanghua Yu, Xintao Wang, Mingdeng Cao, Gen Li, Ying Shan, and Chao Dong.
\newblock Osrt: Omnidirectional image super-resolution with distortion-aware transformer.
\newblock In {\em Proceedings of the IEEE/CVF Conference on Computer Vision and Pattern Recognition}, pages 13283--13292, 2023.

\bibitem{Spherephd}
Yeonkun Lee, Jaeseok Jeong, Jongseob Yun, Wonjune Cho, and Kuk-Jin Yoon.
\newblock Spherephd: Applying cnns on a spherical polyhedron representation of 360deg images.
\newblock In {\em Proceedings of the IEEE/CVF Conference on Computer Vision and Pattern Recognition}, pages 9181--9189, 2019.

\bibitem{32}
Geoffrey Hinton, Oriol Vinyals, and Jeff Dean.
\newblock Distilling the knowledge in a neural network.
\newblock {\em arXiv preprint arXiv:1503.02531}, 2015.

\bibitem{33}
Frederick Tung and Greg Mori.
\newblock Similarity-preserving knowledge distillation.
\newblock In {\em Proceedings of the IEEE/CVF international conference on computer vision}, pages 1365--1374, 2019.

\bibitem{34}
Lucas Beyer, Xiaohua Zhai, Am{\'e}lie Royer, Larisa Markeeva, Rohan Anil, and Alexander Kolesnikov.
\newblock Knowledge distillation: A good teacher is patient and consistent.
\newblock In {\em Proceedings of the IEEE/CVF conference on computer vision and pattern recognition}, pages 10925--10934, 2022.

\bibitem{Focal}
Zhendong Yang, Zhe Li, Xiaohu Jiang, Yuan Gong, Zehuan Yuan, Danpei Zhao, and Chun Yuan.
\newblock Focal and global knowledge distillation for detectors.
\newblock In {\em Proceedings of the IEEE/CVF Conference on Computer Vision and Pattern Recognition}, pages 4643--4652, 2022.

\bibitem{Wavelet}
Linfeng Zhang, Xin Chen, Xiaobing Tu, Pengfei Wan, Ning Xu, and Kaisheng Ma.
\newblock Wavelet knowledge distillation: Towards efficient image-to-image translation.
\newblock In {\em Proceedings of the IEEE/CVF Conference on Computer Vision and Pattern Recognition}, pages 12464--12474, 2022.

\bibitem{Variational}
Sungsoo Ahn, Shell~Xu Hu, Andreas Damianou, Neil~D Lawrence, and Zhenwen Dai.
\newblock Variational information distillation for knowledge transfer.
\newblock In {\em Proceedings of the IEEE/CVF conference on computer vision and pattern recognition}, pages 9163--9171, 2019.

\bibitem{111}
Wenze Hu, Xue Dong, Ning Liu, and Yuanfeng Chen.
\newblock Lumde: Light-weight unsupervised monocular depth estimation via knowledge distillation.
\newblock {\em Applied Sciences}, 12(24):12593, 2022.

\bibitem{222}
Junjie Hu, Chenyou Fan, Hualie Jiang, Xiyue Guo, Yuan Gao, Xiangyong Lu, and Tin~Lun Lam.
\newblock Boosting lightweight depth estimation via knowledge distillation.
\newblock In {\em International Conference on Knowledge Science, Engineering and Management}, pages 27--39. Springer, 2023.

\bibitem{31}
Yiran Wang, Xingyi Li, Min Shi, Ke~Xian, and Zhiguo Cao.
\newblock Knowledge distillation for fast and accurate monocular depth estimation on mobile devices.
\newblock In {\em Proceedings of the IEEE/CVF Conference on Computer Vision and Pattern Recognition}, pages 2457--2465, 2021.

\bibitem{convnet}
Zhuang Liu, Hanzi Mao, Chao-Yuan Wu, Christoph Feichtenhofer, Trevor Darrell, and Saining Xie.
\newblock A convnet for the 2020s.
\newblock {\em Proceedings of the IEEE/CVF Conference on Computer Vision and Pattern Recognition (CVPR)}, 2022.

\bibitem{sup}
Wenzhe Shi, Jose Caballero, Ferenc Huszár, Johannes Totz, Andrew~P. Aitken, Rob Bishop, Daniel Rueckert, and Zehan Wang.
\newblock Real-time single image and video super-resolution using an efficient sub-pixel convolutional neural network.
\newblock In {\em 2016 IEEE Conference on Computer Vision and Pattern Recognition (CVPR)}, pages 1874--1883, 2016.

\bibitem{tracking}
Jianfeng Li, Shigang Li, Tong Chen, and Yiguang Liu.
\newblock Tracking on full-view image for camera motion estimation based on spherical model.
\newblock In {\em 2017 IEEE International Conference on Robotics and Biomimetics (ROBIO)}, pages 569--574. IEEE, 2017.

\bibitem{spherenet}
Benjamin Coors, Alexandru~Paul Condurache, and Andreas Geiger.
\newblock Spherenet: Learning spherical representations for detection and classification in omnidirectional images.
\newblock In {\em Proceedings of the European conference on computer vision (ECCV)}, pages 518--533, 2018.

\bibitem{laina2016deeper}
Iro Laina, Christian Rupprecht, Vasileios Belagiannis, Federico Tombari, and Nassir Navab.
\newblock Deeper depth prediction with fully convolutional residual networks.
\newblock In {\em 3D Vision (3DV), 2016 Fourth International Conference on}, pages 239--248. IEEE, 2016.

\bibitem{sun}
Shuran Song, Fisher Yu, Andy Zeng, Angel~X Chang, Manolis Savva, and Thomas Funkhouser.
\newblock Semantic scene completion from a single depth image.
\newblock In {\em Proceedings of the IEEE conference on computer vision and pattern recognition}, pages 1746--1754, 2017.

\bibitem{sce}
Ankur Handa, Viorica P{\u{a}}tr{\u{a}}ucean, Simon Stent, and Roberto Cipolla.
\newblock Scenenet: An annotated model generator for indoor scene understanding.
\newblock In {\em 2016 IEEE International Conference on Robotics and Automation (ICRA)}, pages 5737--5743. IEEE, 2016.

\end{thebibliography}

\end{document}